%% file: root.tex
\documentclass[letterpaper, 10 pt, conference]{ieeeconf}  

\IEEEoverridecommandlockouts                              

\usepackage{cite}
\usepackage{graphics} 
\usepackage{epsfig} 
\usepackage{times} 
\usepackage{amsmath} 
\usepackage{amssymb}  
\usepackage{amsfonts}
\usepackage{booktabs}
\usepackage{caption}
\usepackage{graphicx}
\usepackage{float} 
\usepackage{algorithm}
\usepackage{algorithmicx}
\usepackage{algpseudocode}
\usepackage{setspace}
\usepackage{subcaption}
\usepackage{xcolor}
\usepackage{bm}
\usepackage{multirow} 
\usepackage{ulem}

\newtheorem{remark}{Remark}

\newcommand{\alg}{CoCoL}

\definecolor{purple}{RGB}{128, 0, 128}

\usepackage{ifthen}
\def\ishidecomments{no}  
\def\authortextcolor{red} 
\newcommand{\jm}[2][]{%
   \ifthenelse{\equal{#1}{}}
   {\textcolor{\authortextcolor}{#2}~\ignorespaces}
      {\ifthenelse{\equal{\ishidecomments}{yes}}
         {\textcolor{\authortextcolor}{#2}~\ignorespaces} 
         {\textcolor{\authortextcolor}{#2}~\textcolor{cyan}{[\textbf{jm}:~#1]}~\ignorespaces}
      }
}

\algrenewcommand\algorithmicindent{1.0em}%
\title{\LARGE \bf
{\alg}: A Communication Efficient Decentralized Collaborative Learning Method for Multi-Robot Systems
}

\newcommand{\btheta}{\bm{\theta}}
\newcommand{\by}{\bm{y}}

\author{
Jiaxi Huang$^\text{1*}$, Yan Huang$^\text{2*}$, Yixian Zhao$^\text{1}$, Wenchao Meng$^\text{1}$, and Jinming Xu$^\text{1}$ 
\thanks{This work has been supported by NSFC under Grants 62088101, 62373323, 62003302.}
\thanks{$^\text{1}$Jiaxi Huang, Yixian Zhao, Wenchao Meng and Jinming Xu are with the College of Control Science and Engineering, Zhejiang University, Hangzhou 310027, China.}
\thanks{
$^\text{2}$Yan Huang is with the Division of Decision and Control Systems, KTH Royal Institute of Technology, Stockholm, Sweden.}
\thanks{{$^*$}Equal Contribution.}
\thanks{Correspondence to {\tt\small jimmyxu@zju.edu.cn} (Jinming Xu). }
}

\renewcommand{\baselinestretch}{0.99}

\begin{document}

\maketitle
\thispagestyle{empty}
\pagestyle{empty}

\begin{abstract}
Collaborative learning enhances the performance and adaptability of multi-robot systems in complex tasks but faces significant challenges due to high communication overhead and data heterogeneity inherent in multi-robot tasks. To this end, we propose {\alg}, a \underline{Co}mmunication efficient decentralized \underline{Co}llaborative \underline{L}earning method tailored for multi-robot systems with heterogeneous local datasets. 
Leveraging a mirror descent framework, {\alg} achieves remarkable communication efficiency with approximate Newton-type updates by capturing the similarity between objective functions of robots, and reduces computational costs through inexact sub-problem solutions. Furthermore, the integration of a gradient tracking scheme ensures its robustness against data heterogeneity.
Experimental results on three representative multi-robot collaborative learning tasks show that the proposed {\alg} can significantly reduce both the number of communication rounds and total bandwidth consumption while maintaining state-of-the-art accuracy.
These benefits are particularly evident in challenging scenarios involving non-IID (non-independent and identically distributed) data distribution, streaming data, and time-varying network topologies.
\end{abstract}

\section{INTRODUCTION}
Multi-robot systems offer the ability to tackle complex tasks through proper collaboration with enhanced efficiency, robustness, and flexibility compared to single-robot systems \cite{rizk2019cooperative}. 
By sharing information, a team of robots can leverage collective knowledge to make more informed decisions and accomplish tasks in a coordinated manner.
Recently, the integration of machine learning techniques into multi-robot systems has led to significant advancements across various domains such as multi-robot SLAM \cite{hu2024cp}, collaborative mapping \cite{deng2024macim}, multi-robot exploration \cite{tan2022deep}, and formation control \cite{bai2021learning}, highlighting the great potential of collaborative learning in large-scale multi-robot applications. 

\begin{figure}[t]
    \centering\includegraphics[width=0.95\linewidth]{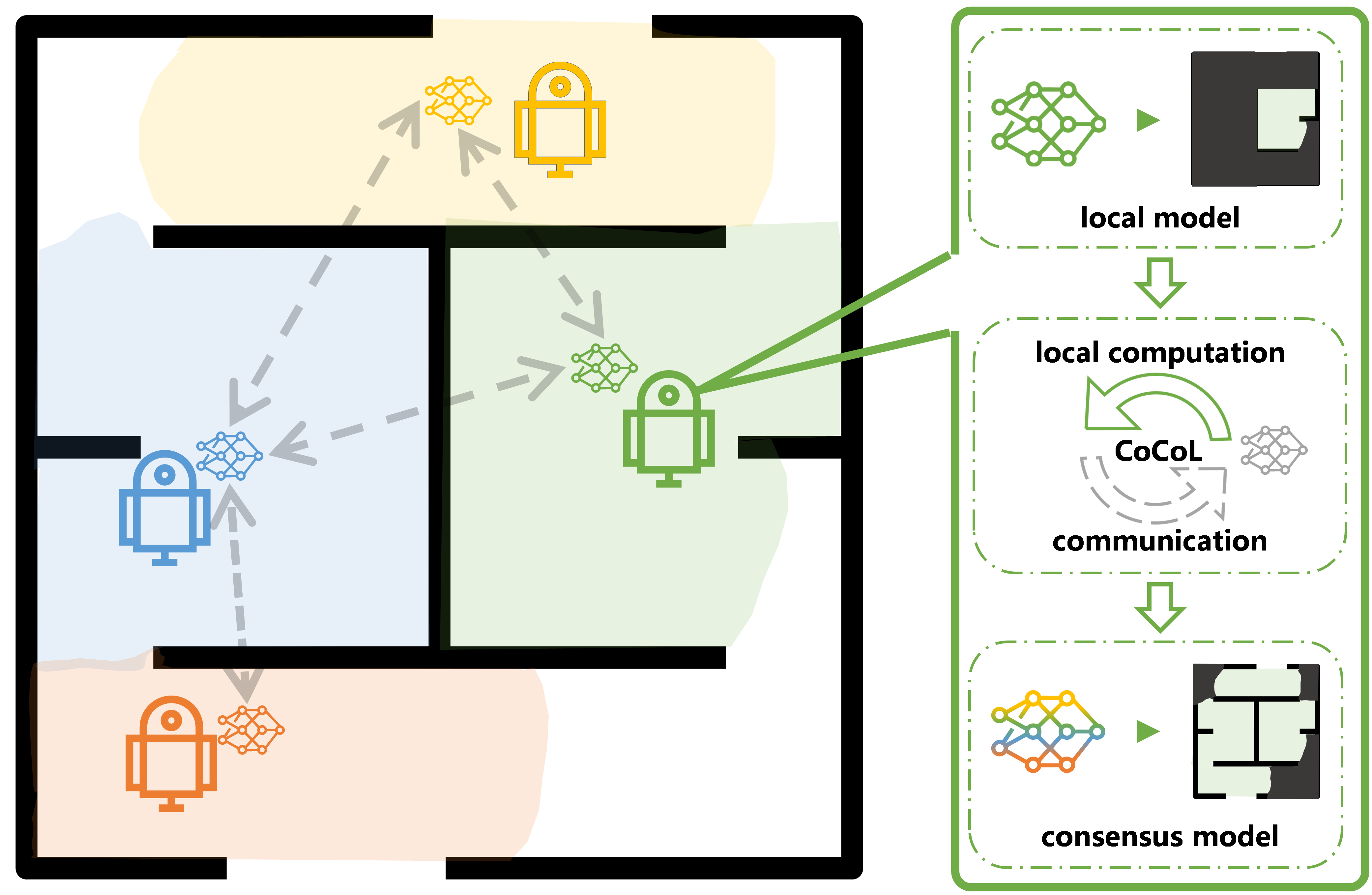} 
    \caption{
    Illustration of a decentralized multi-robot mapping scheme as an example of collaborative learning. Each robot maintains a neural network model for its local map. After several rounds of local computation and communication (grey dashed lines) via collaborative learning, the robots reach a consensus on a global map.
    }
    \label{fig:illustration}
\end{figure}

In collaborative learning, multiple robots collect local data independently and train a shared neural network through inter-robot communication \cite{reisizadeh2019robust}. This learning process can be implemented in either a centralized or decentralized manner. Under a centralized setting, data collected from all robots is aggregated into a leading robot or a central node for computation, with copies of optimized models distributed back to all robots \cite{shorinwa2024distributed}. In contrast, the decentralized scheme (see Fig.~\ref{fig:illustration}) allows each robot to perform on-device computation and exchange information (e.g., model weights, gradients) with neighboring robots, ultimately reaching a consensus on the model parameters maintained by all robots.

However, due to limited onboard resources and varying environmental conditions, decentralized collaborative learning in multi-robot systems faces significant challenges, including high communication overhead and data heterogeneity among robots \cite{chang2020distributed}. For instance, frequent parameter exchanges can cause communication bottlenecks, while the distribution of data collected by each robot may differ due to variations in spatial location, sensors or assigned tasks \cite{cao2023communication}. These characteristics render decentralized algorithms inefficient or even non-convergent. In the past few years, many distributed optimization algorithms, such as gradient tracking \cite{dsgt} and consensus Alternating Direction Method of Multipliers (C-ADMM) \cite{cadmm}, have been widely applied to machine learning tasks, e.g., image classification \cite{lian2017can} and reinforcement learning \cite{zhang2021finite}, with provable efficiency and scalability. 
Recently, Yu et al. \cite{dinno} proposed DiNNO, an ADMM-based collaborative learning algorithm for multi-robot systems, which exhibits the state-of-the-art (SOTA) accuracy on a variety of tasks, underscoring the effectiveness of distributed optimization in such systems. However, existing methods often suffer from relatively slow convergence, resulting in high communication overhead under constrained resources.

In this paper, we propose {\alg}, a communication efficient collaborative learning method for multi-robot systems with heterogeneous data distribution. The contributions of this work are summarized as follows:
\begin{itemize}
    \item We propose {\alg} to solve the collaborative learning problem over networks leveraging the mirror descent framework. {\alg} achieves communication efficient collaborative learning by exploiting the similarity between node objective functions and reduces the computational overhead via inexact sub-problem solutions. Furthermore, the proposed algorithm mitigates the effect of heterogeneous local data distributions by properly incorporating the gradient tracking scheme.
    \item  We conduct comprehensive experiments on three typical collaborative learning tasks for multi-robot systems, demonstrating the effectiveness of {\alg} in reducing communication cost in terms of both the number of communication rounds and total bandwidth, while maintaining state-of-the-art accuracy in various challenging situations, such as non-IID data distribution, streaming data, and time-varying network topologies.
\end{itemize}

\section{Related work}

\subsection{Decentralized Optimization Methods}
Distributed optimization problems have been extensively studied in the past few decades and many efficient algorithms have been proposed for different scenarios \cite{nedic2009distributed, cadmm, di2016next, koloskova2020unified, dsgt, lu2019gnsd}. 
Specifically, in non-IID scenarios, gradient tracking methods \cite{di2016next, dsgt} can asymptotically mitigate the effects of data heterogeneity by incorporating a global gradient estimator. A similar result can be achieved with ADMM-based methods \cite{cadmm}, which employ a primal-dual structure to enforce consensus via dual updates. To reduce computational costs in solving the local subproblem, inexact versions of ADMM-based methods \cite{IC-ADMM, ling2015dlm} have been proposed using a single gradient step for the local update. Despite these improvements in handling data heterogeneity and lowering computational costs, communication overhead remains a significant challenge.

Recent advances in communication efficient distributed optimization reduce communication overhead through two primary approaches: minimizing data volume per communication round and decreasing the total number of communication rounds for reaching certain accuracy. For data volume reduction, quantization schemes \cite{alistarh2017qsgd,zhao2022beer} have proven effective at compressing model and gradient information during transmission. To reduce communication frequency, methods such as \cite{k-gt,nguyen2023performance} introduce multiple steps of local updates between communication steps.
A parallel line of research has emerged focusing on accelerating optimization convergence through implicit second-order information within a mirror-descent framework. For instance, Shamir et al. \cite{DANE} proposed DANE, a centralized approximate Newton-type method that captures the similarity (see \cite[Definition~2.1]{sonata}) between the objective functions of robots. When local objectives exhibit sufficient similarity, DANE and its inexact variant \cite{AIDE} achieve enhanced convergence rates. In decentralized settings, SONATA\cite{sonata} and Network-DANE \cite{li2020communication} incorporate mirror descent updates and gradient tracking to extend DANE to account for decentralized networks. However, the aforementioned mirror descent-based methods focus exclusively on deterministic settings, limiting their applicability to collaborative learning problems in multi-robot tasks. Additionally, some distributed quasi-Newton methods explicitly estimate the aggregate Hessians by introducing auxiliary variables \cite{shorinwa2024quasi,eisen2019primal}, but these approaches require extra transmission overhead per communication round.

\subsection{Collaborative Learning in Multi-Robot Systems}
Applying distributed optimization methods to collaborative learning tasks in multi-robot systems is a promising way with several practical applications. For instance, Hu et al. \cite{hu2024cp} proposed a collaborative SLAM algorithm based on federated learning\cite{mcmahan2017communication}. Another popular task is multi-agent reinforcement learning (MARL), initially explored through the centralized training with decentralized execution (CTDE) paradigm \cite{yu2022surprising,lowe2017multi,10611573}. Subsequently, distributed methods, such as gradient tracking \cite{dsgt} have been integrated with policy gradient techniques for fully decentralized MARL \cite{zhang2021finite,jiang2022mdpgt,chen2024decentralized}, achieving improved convergence rates. Different from these studies tailored for specific applications, DiNNO \cite{dinno}, instead, proposed a more versatile collaborative learning method applicable to a broad range of multi-robot tasks, building on a consensus ADMM (C-ADMM) framework that adopts an approximate solution to the primal problem. Although DiNNO achieves SOTA accuracy on several tasks, 
it potentially requires more communication rounds to reach consensus compared to gradient tracking methods (see Fig. \ref{fig:loss}). More importantly, the optimization methods used in the above applications rely solely on first-order gradient information. It remains unclear if one can leverage higher-order information for collaborative learning to accelerate convergence, thus improving communication efficiency.

\input{body/problem}
\input{body/algorithm}
\section{Experiments}\label{sec:ex}

We evaluate {\alg} across three distinct multi-robot collaborative learning tasks: image classification, neural implicit mapping, and multi-agent reinforcement learning, which are also considered in \cite{dinno}. To assess the robustness of the algorithms, we also incorporate challenging scenarios such as non-IID data distribution, streaming data, and time-varying network topologies in certain tasks.

\subsection{Experimental Setup}

\subsubsection{Baselines}
We compare our method against state-of-the-art methods, including DiNNO \cite{dinno}, the stochastic gradient tracking method DSGT \cite{dsgt}, and the gradient tracking with local updates K-GT \cite{k-gt}. For a comprehensive comparison, we also include centralized results where both data collection and training are conducted on a centralized node.

\subsubsection{Metrics}
To demonstrate the communication efficiency, we compare performance metrics of each algorithm across all tasks with respect to both the number of communication rounds and the total bandwidth consumption to reach certain accuracy. The total bandwidth consumption is calculated by summing the communication volume of each node across all communication rounds.
Compared to DiNNO, although our proposed gradient-tracking based {\alg} algorithm doubles the data transmission volume per communication round, our experiments confirm that it achieves lower overall communication costs while maintaining SOTA accuracy. 
\subsubsection{Implementation Details}
To ensure a fair comparison, we replicate the task settings, network architectures, and hyperparameters from \cite{dinno}. Since all methods, except DSGT, rely on an inner loop for local optimization, we apply the same number of local steps $T$ for each method. For the image classification task, we set $T=2$ and $\mu=0.1$, while for the other tasks, we set $T=5$ and $\mu=0.001$. Generally, the greater the level of similarity between robots' loss functions, the smaller the value of $\mu$ can be for achieving fast convergence \cite{DANE}. To guarantee the validity of our results, we conduct 10 independent trials and report the average result for each task.

\subsection{Image Classification}
In this task, we explore the image classification problem in a multi-robot setting, where the dataset is split across robots. We use the MNIST dataset \cite{mnist} and employ a 4-layer convolutional neural network as the model architecture. The distribution of the dataset among the robots significantly impacts the convergence rate of the algorithms. To evaluate robustness against data heterogeneity, we consider a highly heterogeneous data distribution where 10 robots, each assigned a unique digit class, communicate within a ring graph topology. All robots use the same validation set.
 
\begin{figure}[!t]
    \includegraphics[width=\linewidth]{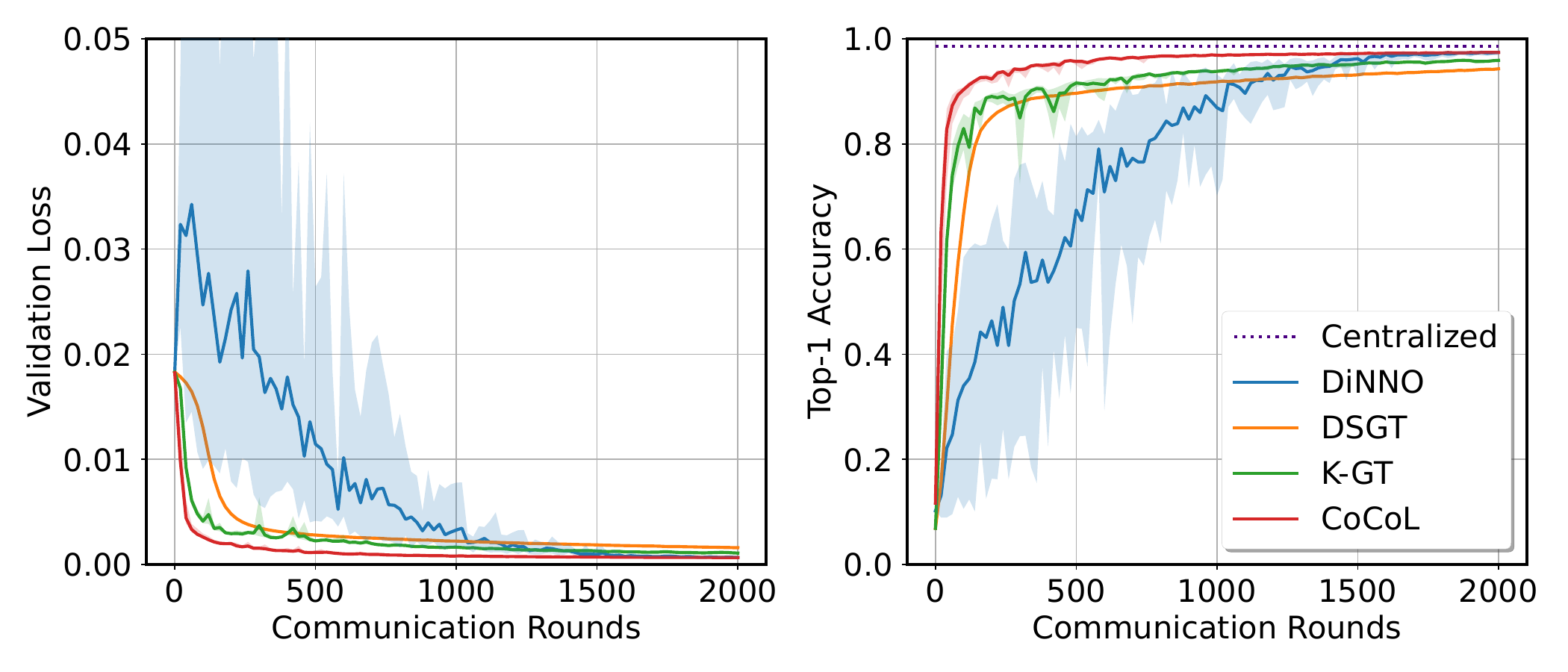} 
    \centering
    \caption{Comparison on validation loss and Top-1 accuracy of the neural network models on each robot under a ring topology. Solid lines depict the average results across all robots, while the shaded areas show the range between the best and worst results at each communication round, which also indicates the consensus error among nodes.} 
    \label{fig:loss}
\end{figure}

Fig. \ref{fig:loss} illustrates the average validation loss and Top-1 accuracy for each aforementioned method, with shaded regions denoting the level of consensus among robots. The smaller the shaded area, the higher consistency across network models will be. It follows that {\alg} not only achieves superior convergence performance despite data heterogeneity but also attains the best final accuracy, matching that of centralized training. Although DiNNO also achieves competitive final accuracy, it necessitates significantly more communication rounds and exhibits considerable oscillation during training, along with higher inconsistency between robots compared to other methods. In contrast, K-GT and DSGT present more stable learning curves, with K-GT slightly outperforming DSGT due to its local update strategy.

\begin{table}[bth]
\renewcommand{\arraystretch}{1.1}
\caption{Comparison on total bandwidth consumption (GB) to reach 95\% Top-1 accuracy.}
\label{table:scale}
\centering
\small
\begin{tabular}{lcccc}
\toprule
\multirow{2}{*}{\textbf{Methods}} & \multicolumn{4}{c}{\textbf{Number of robots}}  \\
\cmidrule(l){2-5}
& 10 & 20 & 50 & 100 \\
\midrule
DSGT \cite{dsgt}      & 7.55 & 23.83 &	77.33 &	234.36  \\
K-GT \cite{k-gt}      & 4.08 & 12.69 & 	41.77  & 124.80    \\
DiNNO \cite{dinno}    & 1.16 & 3.50  & 	13.72 	& 40.27  \\
{\alg} (ours)         & \textbf{1.05} & \textbf{2.36}  & \textbf{9.55}   & 	\textbf{18.87}   \\

\bottomrule
\end{tabular}
\end{table}
To further assess the scalability of different methods, we examine their performance as the number of robots increases, as detailed in TABLE \ref{table:scale}. The communication graph is generated with an algebraic connectivity of 1$\pm 0.01$, and we introduce some degree of data heterogeneity by limiting each robot to access at most two digit classes. TABLE \ref{table:scale} presents the total bandwidth consumption to reach 95\% accuracy. It is evident that {\alg} consistently outperforms the other methods, reducing the total communication volume by 9.1\% to 53.2\% compared to DiNNO across varying numbers of robots. Notably, this advantage becomes more pronounced as the number of robots increases, highlighting its effectiveness and scalability for large-scale applications.



\begin{figure}[!h]
    \centering
    \includegraphics[width=\linewidth]{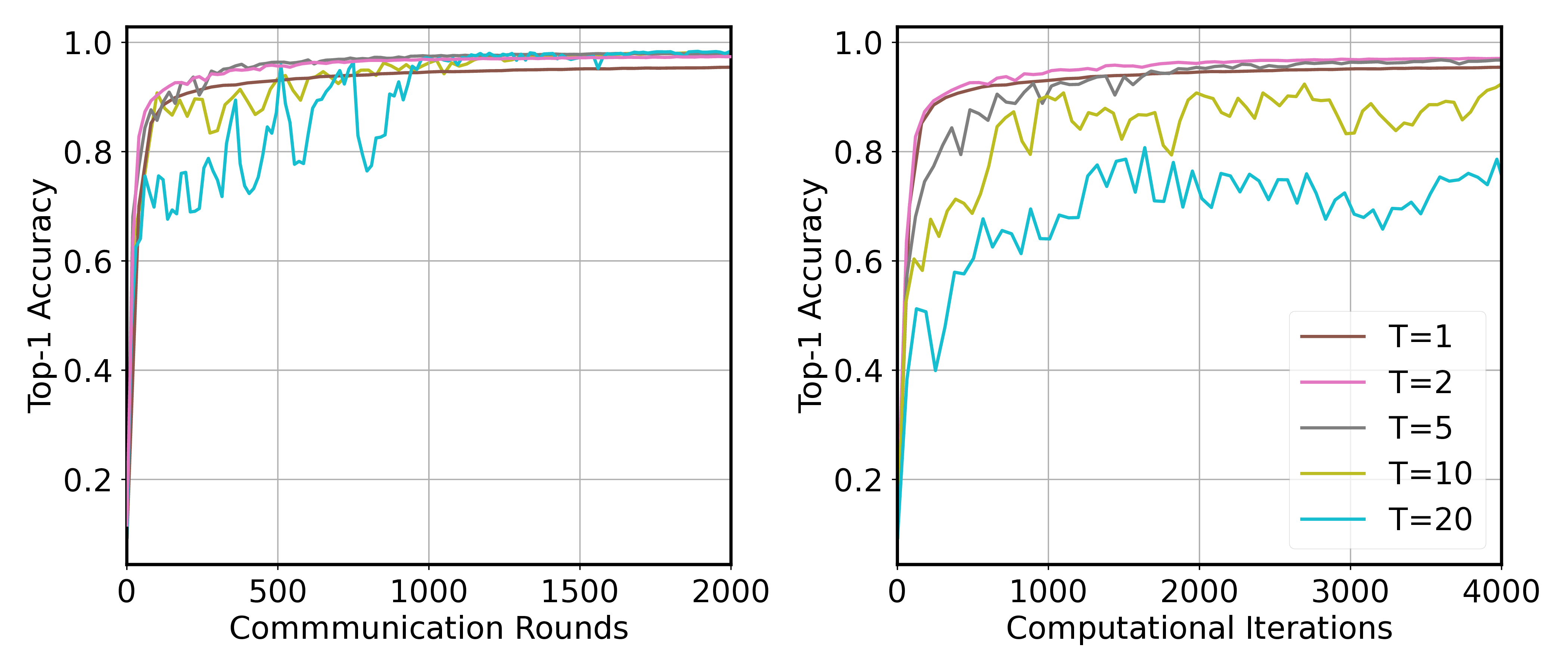}
    \caption{Top-1 accuracy versus number of communication rounds (left) and computational iterations (right) for different local steps. $T=2$ presents the best trade-off between computational efficiency and communication overhead.}
    \label{fig:local_step}
\end{figure}

Moreover, we evaluate the convergence behavior of {\alg} under different local steps $T$, as illustrated in Fig. \ref{fig:local_step}. Intuitively, a large local step can resemble the behavior of SONATA, where the local sub-problem should be solved exactly. In contrast to the expectation that increasing $T$ would lead to fewer communication rounds, the left panel of Fig. \ref{fig:local_step} reveals that it achieves greater communication efficiency with $T=2$ and $T=5$ in the context of this particular problem and data distribution. Additionally, the right panel of Fig. \ref{fig:local_step} evaluates the accuracy with respect to computational iterations, measured as the number of forward/backward passes on a single robot. It is evident that CoCoL achieves the highest accuracy using the least computational effort with $T=2$. This result is particularly significant for practical multi-robot systems, where computational resources and communication conditions can vary. The number of local steps is crucial in balancing computation and communication efficiency. In fact, the proper choice of local steps depends on the similarity between local Hessians. In particular, when local Hessians exhibit higher similarity, more rounds of local updates can be performed, significantly reducing communication overhead without much additional computational cost. Note that there exists an optimized number of local iterations, beyond which computational efficiency degrades. Similar observations have also been found in \cite{10852183}.

\subsection{Neural Implicit Mapping}
\begin{figure}[t]
    \includegraphics[width=\linewidth]{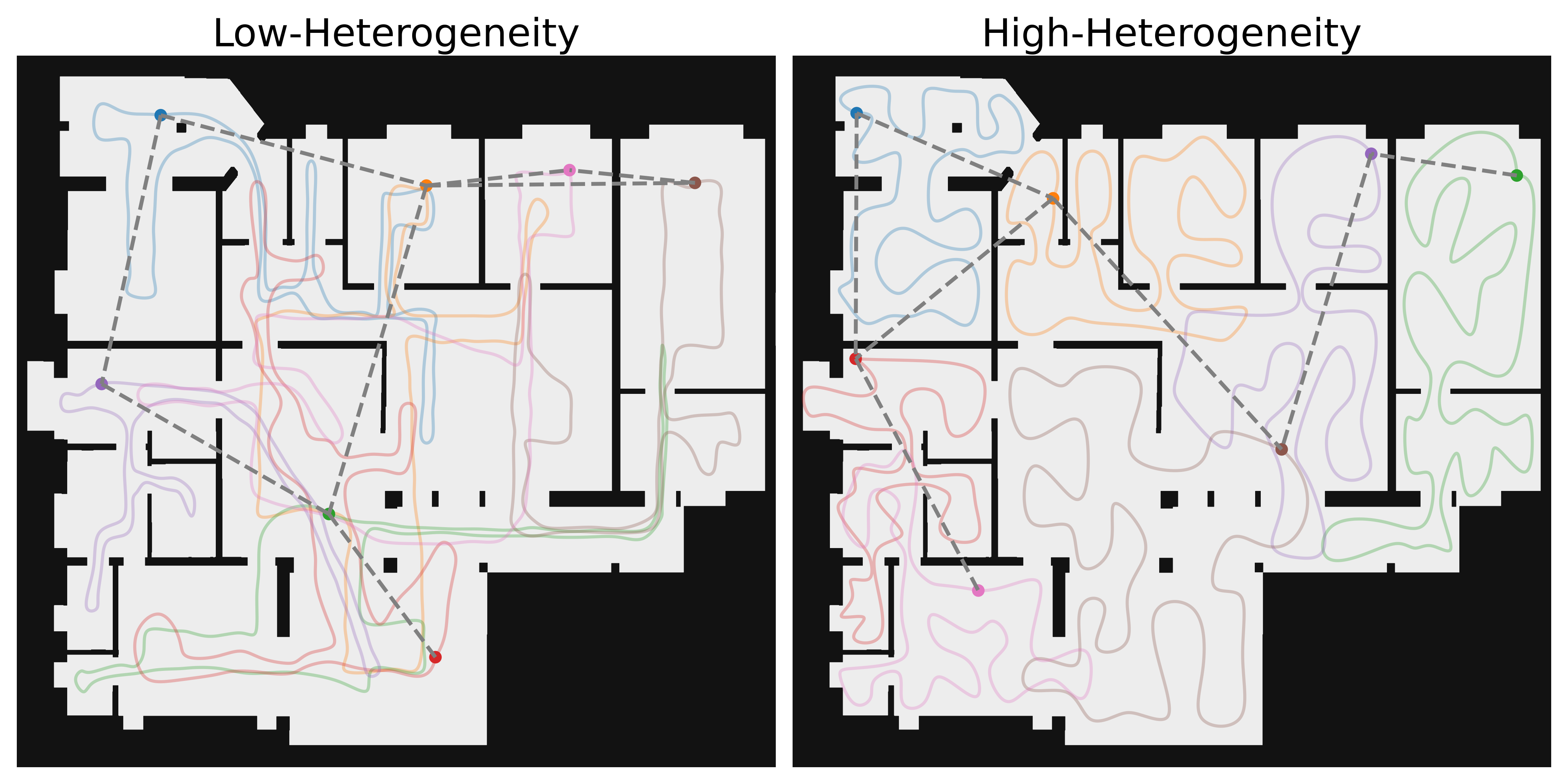} 
    \centering
    \caption{Ground truth environment with two trajectory scenarios: a low-heterogeneity scenario on the left and a high-heterogeneity scenario on the right, with minimal overlap between trajectories. The gray dashed lines represent the time-varying communication graph among robots.}
    \label{fig:two path}
\end{figure}

In this section, we evaluate {\alg} on the multi-robot two-dimensional neural implicit mapping task. 
Neural implicit mapping, such as NeRF \cite{mildenhall2021nerf}, uses neural networks to represent continuous spatial functions. 
We employ a Multi-Layer Perceptron (MLP) with sinusoidal activation \cite{tancik2020fourier} to model a 2D floor plan occupancy map. Given spatial coordinate $(x,y)$ as input, the network predicts a density value $\sigma\in(0,1)$ to represent the occupancy status at that location. We use the dataset in \cite{dinno} that includes trajectory points and corresponding lidar scans for 7 robots. Each robot is equipped with its global location and optimizes its local network using 400 latest lidar scans, emulating real-time data flow. The validation set consists of unseen lidar scans from randomly sampled positions on the map, excluding areas inside walls. 
As depicted in Fig. \ref{fig:two path}, we test two trajectory scenarios with varying degrees of data heterogeneity. 
The first scenario with relatively low data heterogeneity is characterized by substantial overlap between trajectories, resulting in similar point cloud scans and shared observations among robots.
The second scenario with high heterogeneity features minimal overlap between paths, which increases the likelihood that robots may get trapped in local optima and struggle to achieve global consensus due to limited shared observations. Note that the communication graph in both scenarios is time-varying and constructed based on robots' communication range to mimic real-world conditions.
\begin{table}[!bht]
\renewcommand{\arraystretch}{1.1}
\caption{\textbf{Quantitative comparison of mapping results.}}
\label{table:metric}
\centering
\small
\begin{tabular}{lccccccc}
\toprule
\multirow{2}{*}{Method} & \multicolumn{2}{c}{Val. Loss $\downarrow$} & \multicolumn{2}{c}{MSE $\downarrow$} & \multicolumn{2}{c}{SSIM $\uparrow$} \\
\cmidrule(lr){2-3} \cmidrule(lr){4-5} \cmidrule(lr){6-7}
& Low & High & Low & High & Low & High \\
\midrule
Centralized & 0.89 & 0.88 & 0.012 & 0.016 & 0.94 & 0.90 \\
\midrule
DSGT \cite{dsgt}      & 4.60 & 4.79  & 0.295 & 0.297 & 0.10 & 0.07 \\
K-GT \cite{k-gt}      & 4.78 & 5.28 & 0.303 & 0.298 & 0.09 & 0.07 \\
DiNNO \cite{dinno}    & 2.79 & 2.94 & 0.125 & 0.127 & 0.66 & 0.64 \\
{\alg} (ours) & \textbf{1.13} & \textbf{1.36} & \textbf{0.034} & \textbf{0.040} & \textbf{0.88} & \textbf{0.85} \\
\bottomrule
\end{tabular}
\end{table}
\begin{figure}[th]
    \includegraphics[width=\linewidth]{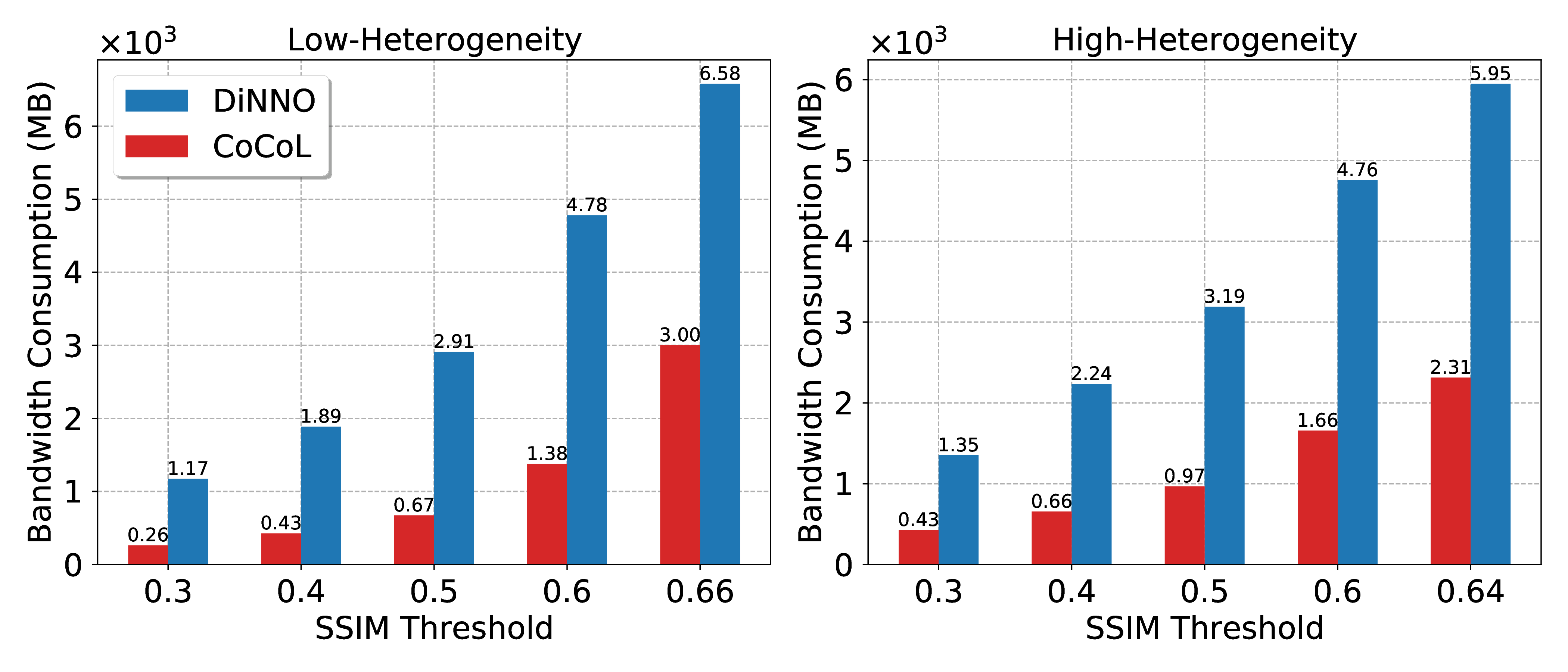} 
    \centering
    \caption{Comparison on total bandwidth consumed across different SSIM thresholds under two scenarios.}
    \label{fig:ssim}
\end{figure}

\begin{figure*}[thpb]
    \includegraphics[width=\linewidth]{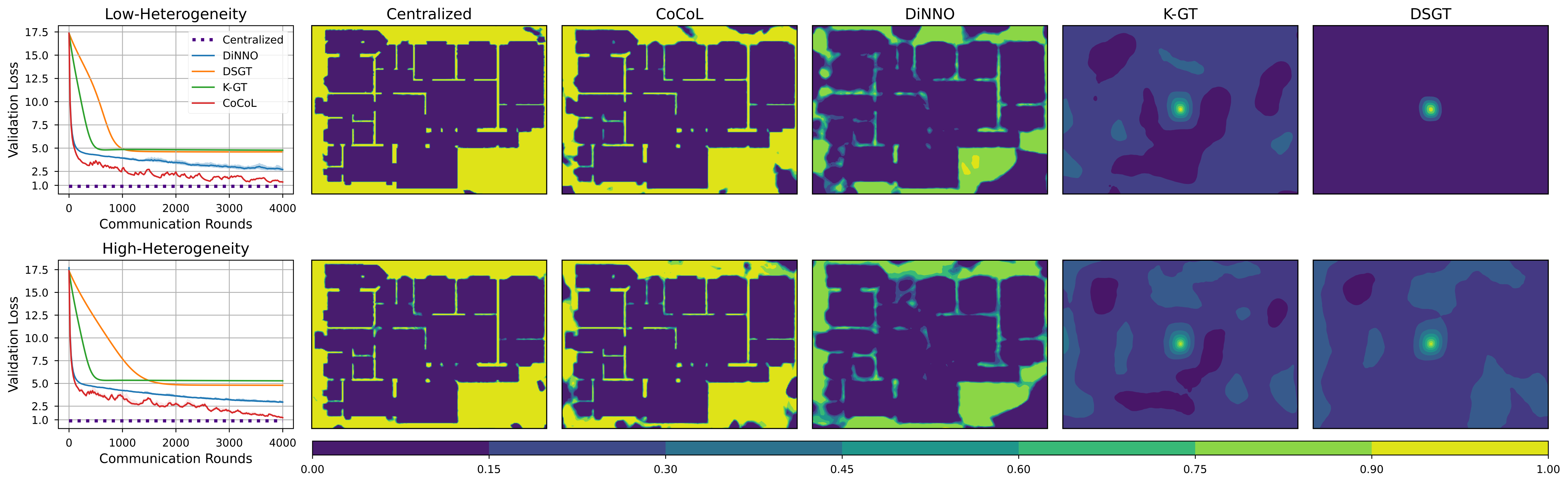} 
    \centering
    \caption{The first column plots the mean validation loss over communication rounds, with shaded areas representing consensus error. The following columns display the reconstructed maps for both centralized and decentralized methods, generated by feeding grid mesh points from the map into the optimized network model of an arbitrary robot. Our method yields the most accurate and detailed reconstructions, comparable to those achieved by the centralized approach.}
    \label{fig:compare}
\end{figure*}

Fig. \ref{fig:compare} presents the validation loss and reconstructed maps for a randomly selected robot in both scenarios. It follows that {\alg} consistently outperforms alternative methods in terms of both convergence rate and mapping quality across two scenarios, comparable to centralized training performance. Notably, {\alg} reaches the final validation loss of DiNNO in nearly 25\% of the communication rounds. These substantial improvements in communication complexity do not compromise mapping quality. Even in the high-heterogeneity scenario where robots are confined to separate rooms with limited joint observation, {\alg} enables each robot to reconstruct the global map accurately. While DiNNO captures the overall floor plan structure, it produces noticeably blurred details and lower predicted density values in occupied areas. K-GT and DSGT both fail to produce coherent maps under the dynamic conditions of streaming data and time-varying graphs. 


Treating the occupancy map as an image, we compute the Mean Squared Error (MSE) and Structural Similarity Index (SSIM) \cite{wang2004image} between predicted density values and the ground truth. It follows from Table \ref{table:metric} that {\alg} improves the quality of the reconstructed map by 33\% in terms of SSIM compared to DiNNO, closely matching the performance of the centralized approach. The slight discrepancy in MSE and SSIM between the centralized method and {\alg} can be attributed to the presence of holes within walls, as these metrics are evaluated across the entire map.

We also measure the total bandwidth required to reach a certain SSIM threshold by CoCoL and DiNNO in Fig \ref{fig:ssim}. In both low and high heterogeneity scenarios, {\alg} consistently achieves superior communication efficiency compared to DiNNO, saving 54.4\% to 77.8\% of the total communication volume across different SSIM thresholds.

\subsection{Decentralized Multi-Agent Reinforcement Learning}

To test the generality of the method, in the third experiment, we evaluate a decentralized multi-agent reinforcement learning task, the multi-robot predator-prey problem in \cite{lowe2017multi}. 
This task is particularly challenging due to its decentralized nature in both training and execution, as well as the non-stationarity of the environment, where each agent relies on local observations.
The objective is for 3 slower robots to cooperatively pursue a faster adversary within a bounded region containing eight landmarks. During training, agents that fail to coordinate effectively would receive lower reward, while successful collisions with the adversary result in a joint reward. Meanwhile, the adversary uses a heuristic policy to evade capture by moving away from the closest robot and avoiding boundaries.

To address this problem, we implement a decentralized version of PPO by integrating {\alg} with Proximal Policy Optimization (PPO) as also used in \cite{dinno}, aiming for consistent policy updates. We apply this integration to all distributed optimization methods considered in this experiment. We note that the parameters for both the actor and critic networks are updated separately in each iteration. The communication graph is fully connected, and MLPs with ReLU activations are used for both networks' architectures.

\begin{figure}[tpb]
    \includegraphics[width=\linewidth]{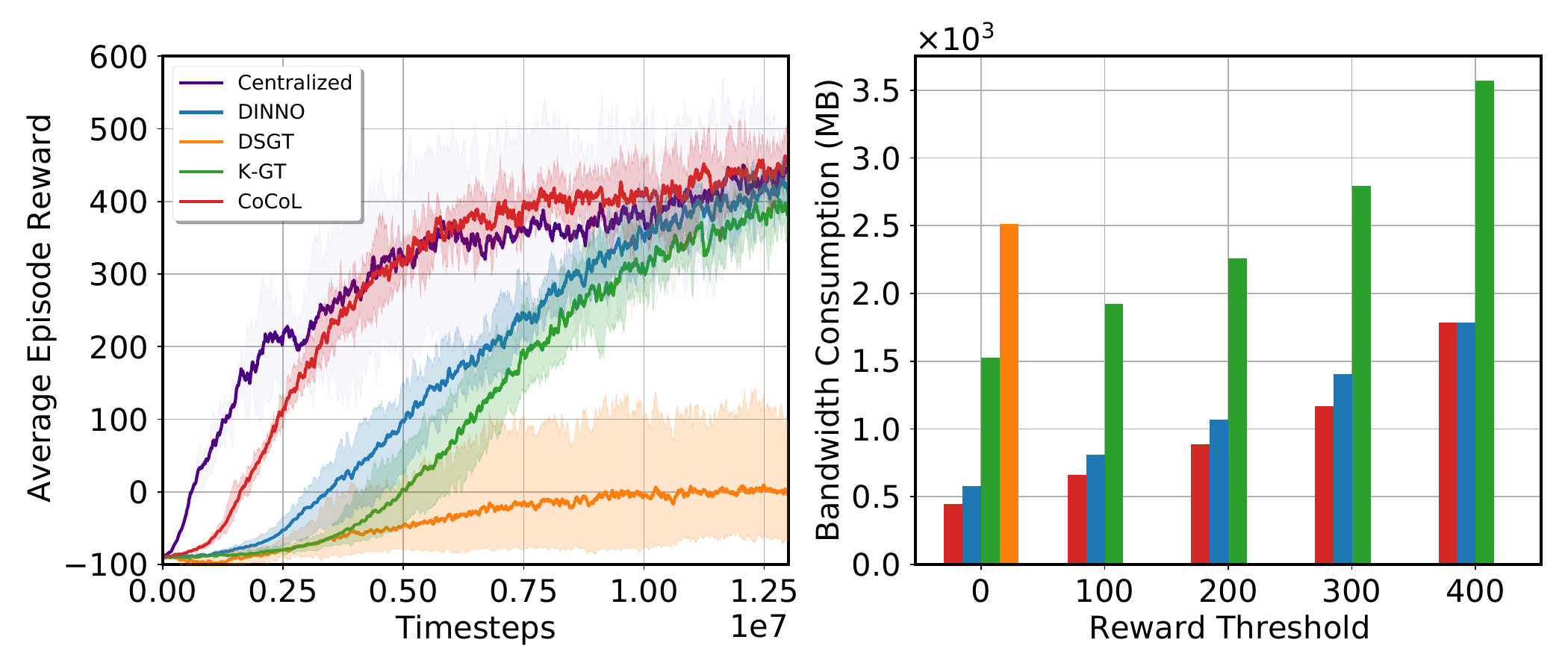} 
    \centering
    \caption{Left: The average episodic reward versus environment time steps for all methods. Solid lines represent the mean results based on 10 random seeds. The shaded areas indicate the range between upper and lower bounds across seeds. Right: Comparison on total bandwidth consumed across different reward thresholds.}
    \label{fig:reward}
\end{figure}

We evaluate each method using 10 random seeds and plot the mean and range of the average episodic reward in Fig. \ref{fig:reward}, along with the total bandwidth consumption for reaching different reward threshold. Both {\alg} and DiNNO achieve nearly the same final average reward as the centralized approach. However, {\alg} converges significantly faster, matching the reward growth rate of the centralized approach.
K-GT achieves slightly inferior results, while DSGT performs the worst, exhibiting significant variability across different random seeds. In terms of total bandwidth consumption in the right panel of Fig.~\ref{fig:reward}, {\alg} demonstrates comparable bandwidth utilization to DiNNO in this task, which may be attributed to the simple fully connected communication topology.

\section{CONCLUSIONS}
In this paper, we have proposed {\alg}, a communication efficient decentralized collaborative learning method designed for multi-robot systems. 
Our approach integrates the strength of the mirror descent framework with gradient tracking, thereby effectively addressing the challenges of high communication overhead and data heterogeneity that are common in multi-robot tasks. Experiments across diverse collaborative learning tasks show that {\alg} significantly reduces communication overhead while maintaining the state-of-art accuracy. The proposed {\alg} is particularly useful in scenarios where onboard resources are limited. Future work will focus on further enhancing communication efficiency by employing certain quantization mechanism. Also, it is of great interest and importance to explore the potential application of {\alg} in more sophisticated learning tasks such as collaborative SLAM.
\normalem
\bibliographystyle{IEEEtran}
\bibliography{IEEEabrv,refs}

\end{document}

%% file: body/problem.tex
\section{Problem Formulation}
Consider a multi-robot system in a connected communication graph $\mathcal{G} = (\mathcal{V}, \mathcal{E})$, where $\mathcal{V}=\{1,..,N\}$ denotes the set of robots and $\mathcal{E}\subseteq\mathcal{V}\times\mathcal{V}$ represents the undirected communication links between robots. 
Subsequently, we introduce a weight matrix $\bm{W}\triangleq \left(w_{ij}\right)^N_{i,j=1}$ to aggregate information among neighboring robots in the graph $\mathcal{G}$. To guarantee consensus among robots, we assume that the weight matrix $\bm{W}$ induced by $\mathcal{G}$ is doubly stochastic, i.e., 
\begin{gather*}    
\bm{W}=
\left\{
\begin{aligned}
    &w_{ij} > 0, &&i=j \text{ or }(i,j) \in \mathcal{E} \\
    &w_{ij} = 0, &&\text{otherwise}
\end{aligned}
\right.\\
\bm{W}^\top\mathbf{1}=\mathbf{1}\quad \text{and} \quad\bm{W}\mathbf{1}=\mathbf{1},
\end{gather*}
where $\mathbf{1}\in \mathbb{R}^N$ is the all-one vector. This weight matrix can be easily determined based on the Metropolis-Hastings protocol \cite{xiao2006distributed} for undirected graphs. 

The entire multi-robot system seeks to complete some deep learning tasks collaboratively. Each robot $i\in \mathcal{V}$ employs the same neural network architectures and accumulates its local data sets $\mathcal{D}_i$. In collaborative learning problems, the goal is to find the best model parameters $\btheta \in \mathbb{R}^d$ that minimizes the following problem:
\begin{align}
\underset{\bm{\theta} \in \mathbb{R}^d}{\text{min}} \quad \frac{1}{N}\sum_{i \in \mathcal{V}} \mathcal{L}_i
(\bm{\theta}; \mathcal{D}_i), \label{eq:joint_loss}
\end{align}
where $\mathcal{L}_i(\cdot)$ is  the local loss function of robot $i$. To solve this problem in a decentralized manner, we introduce a local copy of the model parameters $\bm{\theta}_i\in \mathbb{R}^d$ for each robot $i$. As a result, Problem (\ref{eq:joint_loss}) transfers to its equivalent form:
\begin{equation}\label{Prob_2}
\begin{aligned}
    \underset{\{\bm{\theta}_i,\ \forall i\in \mathcal{V}\}}{\text{min}} \quad \frac{1}{N}\sum_{i \in \mathcal{V}} \mathcal{L}_i(\bm{\theta}_i; \mathcal{D}_i)\\
    \text{s.t.} \quad \bm{\theta}_i=\bm{\theta}_j \quad \forall (i, j) \in \mathcal{E}.
\end{aligned}
\end{equation}

Compared to centralized settings \cite{shorinwa2024distributed}, the challenge for decentralized collaborative learning is that each robot $i$ can only access its local loss function and process its local dataset along with information from its neighbors. A general decentralized collaborative learning usually consists of parallel local computation steps for each robot and a communication step to exchange messages such as current model parameters between neighbors to reach a consensus on decision variables (see Fig.~\ref{fig:illustration}). The proper utilization of exchanged messages during local processing is crucial for algorithm design.

%% file: body/algorithm.tex
\section{Algorithm Design}
This section begins with essential preliminaries, followed by the introduction of our proposed method.
\subsection{Preliminaries}\label{Sec_priliminary}

\textbf{Mirror descent method.} It has been demonstrated that the mirror descent method can be used to improve the efficiency of distributed optimization \cite{sonata}. To see this, we recall the general form of this method with $N$ clients at iteration $k$:
\begin{equation}\label{Eq_mirror_descent_1}
\begin{aligned}
\boldsymbol{\theta }_{i}^{k+1}=\underset{\boldsymbol{\theta }_i}{\mathrm{arg}\min}\left\{ \left< \boldsymbol{\theta }_i,\nabla f\left( \bar{\boldsymbol{\theta}}^k \right) \right> +\frac{1}{\eta}B_{\varphi}\left( \boldsymbol{\theta }_i,\bar{\boldsymbol{\theta}}^k \right) \right\} ,
\end{aligned}
\end{equation}
where $f\left( \boldsymbol{\theta } \right) =1/N\sum_{i=1}^N{f_i\left( \boldsymbol{\theta} \right)}$ with $f_i$ denoting the local convex and differentiable objective function of node $i$, $\boldsymbol{\bar{\theta}}^k=1/N\sum_{i=1}^N{\boldsymbol{\theta }_i^{k}}$, $\eta > 0$ and $B_{\varphi}\left( \boldsymbol{\theta }_i,\boldsymbol{\bar{\theta}}^k \right) $ denotes the Bregman divergence as below:
\[
B_{\varphi}\left( \boldsymbol{\theta }_i,\boldsymbol{\bar{\theta}}^k \right) =\varphi \left( \boldsymbol{\theta }_i \right) -\varphi \left( \boldsymbol{\bar{\theta}}^k \right) -\left< \boldsymbol{\theta }_i-\boldsymbol{\bar{\theta}}^k,\nabla \varphi \left( \boldsymbol{\bar{\theta}}^k \right) \right> ,
\]
where $\varphi \left( \cdot \right) $ is a continuously differentiable and convex function. Setting the reference function $\varphi \left( \boldsymbol{\theta } \right) =f_i\left( \boldsymbol{\theta } \right) +\frac{\mu}{2}\left\| \boldsymbol{\theta } \right\| ^2$ for each node $i$, where $\mu$ is a regularization parameter, the mirror descent method can be reformulated as:
\begin{equation}\label{Eq_mirror_descent_2}
\begin{aligned}
\boldsymbol{\theta }_{i}^{k+1}=\mathrm{arg}\underset{\boldsymbol{\theta }_i}{\min}&\{f_i\left( \boldsymbol{\theta }_i \right) +\left< \boldsymbol{\theta }_i,\eta \nabla f\left( \bar{\boldsymbol{\theta}}^k \right) -\nabla f_i\left( \bar{\boldsymbol{\theta}}^k \right) \right> 
\\
&+\frac{\mu}{2}\left\| \boldsymbol{\theta }_i-\boldsymbol{\bar{\theta}}^k \right\| ^2 \}.
\end{aligned}
\end{equation}
This formulation exactly recovers the DANE algorithm \cite{DANE}, which is a popular communication efficient approximate Newton method with master-slave architectures. In particular, consider the case that the objective function $f_i$ of each node $i$ takes the quadratic form $f_i\left( \boldsymbol{\theta } \right) =\frac{1}{2}\boldsymbol{\theta }^{\top}\boldsymbol{H}_i\boldsymbol{\theta}+\bm{b}_i^\top\bm{\theta}+\bm c_i$ where the Hessian matrix $\boldsymbol{H}_i$ is a fixed symmetric and positive semi-definite matrix. Then, the mirror descent method becomes
\begin{equation}\label{Eq_mirror_descent_similarity}
\begin{aligned}
\boldsymbol{\theta }_{i}^{k+1}&=\bar{\boldsymbol{\theta}}^k-\eta (\boldsymbol{H}_i+\mu \boldsymbol{I}_d)^{-1}\nabla f\left( \bar{\boldsymbol{\theta}}^k \right) 
\\
&=\bar{\boldsymbol{\theta}}^k-\underset{\mathrm{Similarity}}{\eta \underbrace{(\boldsymbol{H}_i+\mu \boldsymbol{I}_d)^{-1}\bar{\boldsymbol{H}}}}\underset{\mathrm{Newton}\,\, \mathrm{method}}{\underbrace{\bar{\boldsymbol{H}}^{-1}\nabla f\left( \bar{\boldsymbol{\theta}}^k \right) }},
\end{aligned}
\end{equation}
where $\bar{\boldsymbol{H}}=1/N\sum\nolimits_{j=1}^N{\boldsymbol{H}_j}$.

The key insight here is to perform an approximate Newton-type update by capturing the similarity between local Hessians $\boldsymbol{H}_i$ and their global mean $\bar{\boldsymbol{H}}$, without the need for explicit Hessian computations. If the local Hessians $\boldsymbol{H}_i$ at each node are sufficiently similar, to the point where the similarity term closely approximates an identity matrix, i.e., $\| \boldsymbol{I}_d-\left( \boldsymbol{H}_i+\mu \boldsymbol{I}_d \right) ^{-1}\bar{\boldsymbol{H}} \| \rightarrow \mu /\left( \lambda +\mu \right)$ where $\lambda$ is the smallest eigenvalue of $\bar{\boldsymbol{H}}$ \cite{DANE}, the update directions of the nodes become well-aligned and matching that of the Newton method. This alignment significantly enhances the convergence performance of mirror descent methods, which in turn reduces the need for communication.

\textbf{Gradient tracking method.} In decentralized scenarios, since each robot only communicates with its neighbors instead of a centralized server, obtaining the accurate average of the model parameters and global gradient at each iteration $k$ can be challenging. Gradient tracking \cite{xu2015augmented} is a popular decentralized optimization method that can achieve asymptotic consensus of model parameters and track the global gradient via local estimators denoted as $\by_i$. In particular, it initializes $\boldsymbol{y}_{i}^0=\nabla f_i\left( \boldsymbol{\theta }_{i}^0 \right)$ and then iterates as follows:
\begin{equation}\label{Eq_gradient_traking}
\begin{aligned}
\boldsymbol{\theta }_i^{k+1}&=\sum_{j=1}^N{w_{ij}\boldsymbol{\theta }_j^{k}-\gamma}\boldsymbol{y}_i^{k},
\\
\boldsymbol{y}_i^{k+1}&=\sum_{j=1}^N{w_{ij}\boldsymbol{y}_j^{k}+}\nabla f_i\left( \boldsymbol{\theta }_i^{k+1} \right) -\nabla f_i\left( \boldsymbol{\theta }_i^{k} \right) ,
\end{aligned}
\end{equation}
where $\gamma$ is the stepsize for local update. When the model parameters reach a consensus, i.e., $\boldsymbol{\theta }_i^{k}\rightarrow \boldsymbol{\bar{\theta}}^k$ as $k$ goes to infinity, then, for the gradient estimator, we have ${\boldsymbol{y}_i^{k}} \rightarrow \nabla f\left( \boldsymbol{\bar{\theta}}^k \right)$ \cite{di2016next,xu2015augmented}. It has been proven that gradient tracking can mitigate the influence of data heterogeneity across nodes \cite{xu2015augmented, dsgt}, and for nonconvex problems, it converges to a stationary point with fixed stepsize \cite{sonata}.

\subsection{The Proposed Method: {\alg} }

As discussed above, most existing mirror-descent-based methods focus on deterministic optimization problems, such as (Network-) DANE \cite{DANE,li2020communication} and SONATA \cite{sonata}. Moreover, these methods require solving the subproblem \eqref{Eq_mirror_descent_2} exactly, which becomes computationally intractable for many collaborative learning tasks in multi-robot scenarios, especially when dealing with commonly stochastic and non-convex loss functions $\mathcal{L}_i(\boldsymbol{\theta}_i;\mathcal{D}_i)$.

To address this issue, we propose a communication-efficient decentralized collaborative learning method, termed {\alg}, with pseudo-code presented in Algorithm \ref{alg:ours}. This algorithm integrates a stochastic gradient tracking mechanism with mirror descent updates. 
Particularly, {\alg} allows for an inexact solution to the following sub-problem with $\eta=1$ at each iteration $k$:
\begin{equation}\label{Eq_subproblem_3}
\begin{aligned}
\hat{\boldsymbol{\theta}}_{i}^{k}\approx &\mathrm{arg}\underset{\boldsymbol{\theta }_i}{\min}\Big\{ g_i\left( \boldsymbol{\theta }_i;\mathcal{D} _i \right) :=\mathcal{L} _i\left( \boldsymbol{\theta }_i;\mathcal{D} _i \right) 
\\
&+\left. \langle \,\boldsymbol{\theta }_i,\eta\boldsymbol{y}_{i}^{k}-\nabla \mathcal{L} _i(\boldsymbol{\theta }_{i}^{k};\mathcal{D} _i)\rangle \right. \left. +\frac{\mu}{2}\left\| \boldsymbol{\theta }_i-\boldsymbol{\theta }_{i}^{k} \right\| ^2 \right\}.
\end{aligned}
\end{equation}
Specifically, at each round, a finite number of local stochastic gradient updates are performed using Adam \cite{Adam} as the optimizer, with $\boldsymbol{\theta }_{i}^{k}$ as the initial point (c.f., Line 5-9 in Algorithm \ref{alg:ours}).
This significantly improves the computational efficiency while capturing the similarity of the local objectives for efficient optimization as illustrated in \eqref{Eq_mirror_descent_similarity}.

To account for data heterogeneity, we further introduce the gradient tracking scheme in the approximate mirror descent update (c.f., Line 10-12 in Algorithm \ref{alg:ours}), resulting in the following update rule of {\alg} algorithm:
\begin{equation}
\begin{aligned}
\boldsymbol{\theta }_{i}^{k+1}&=\sum_{j=1}^N{w_{ij}\hat{\boldsymbol{\theta}}_{j}^{k}},
\\
\boldsymbol{y}_{i}^{k+1}&=\sum_{j=1}^N{w_{ij}\boldsymbol{y}_{j}^{k}}
+\nabla \mathcal{L} _i(\boldsymbol{\theta }_{i}^{k+1};\mathcal{D} _i)-\nabla \mathcal{L} _i(\boldsymbol{\theta }_{i}^{k};\mathcal{D} _i).
\end{aligned}
\end{equation}
By communicating the local parameters $\btheta_i$ and gradient estimator $\by_i$ with the neighbors of each node over the network,
the proposed {\alg} asymptotically closes the gap between local copies of the model parameters and their global average, and tracks the global average gradient. In so doing, it approximates a centralized mirror descent update in a decentralized manner. This approach, compared to solving ~\eqref{Eq_mirror_descent_2} exactly, makes decentralized optimization feasible for collaborative learning systems while benefiting from fast convergence rates.

\begin{algorithm}[t!]
\caption{{\alg}}
\label{alg:ours}
\begin{algorithmic}[1]
\State \textbf{Input:} $\btheta_{ini}\text{: initial value}, \: \bm{W}\text{: weight matrix}, \: \mu \text{: regulari}$
\Statex $\text{-zer},\:T\text{: number of local steps},\: \gamma\text{: learning rate}$

\State \textbf{Initialize:} $\forall i\in \mathcal{V},\ \btheta_i^0\gets\btheta_{ini},\ \by_i^0\gets\nabla\mathcal{L}_i(\btheta_i^0;\mathcal{D}_i) $ \label{alg:line:initial}
    \For{$k=0 \text{ to } K$} 
        \For{$i\in \mathcal{V}$} \textbf{in parallel}
            \State $\hat{\btheta}^{0}_i=\btheta^k_i$
            \For{$t=0 \text{ to } T$}\label{alg:line:local_start}
            \State obtain random sample $\xi^t_i$ from $\mathcal{D}_i$ and calculate
            \Statex \qquad \quad  the stochastic gradient $\nabla g_i(\hat{\btheta}^t_i;\xi^t_i)$ 
            \State $\hat{\btheta}^{t+1}_i=\text{Adam}(\hat{\btheta}^{t}_i,\gamma,\nabla g_i(\hat{\btheta}^t_i;\xi^t_i))$ \label{alg:line:local_end}
            \EndFor
        \State Send $\hat{\btheta}_i^{T} \text{ and } \by_i^k$ to neighbors $\mathcal{N}_i$ 
        \State $\btheta_i^{k+1}=\sum_{j=1}^{N}w_{ij}\hat{\btheta}_j^{T}$\label{alg:line:start}
        \State $\by_i^{k+1}=\sum_{j=1}^{N}w_{ij}\by_j^k$ \label{alg:line:end}
        \Statex\qquad\ $+\nabla\mathcal{L}_i(\bm{\theta}_i^{k+1};\mathcal{D}_i)-\nabla\mathcal{L}_i(\bm{\theta}_i^k ;  \mathcal{D}_i)$
        \EndFor
            
    \EndFor  
    \State \Return{$\{\btheta_i^{K}\}_{i \in \mathcal{V}}$}
\end{algorithmic}
\end{algorithm}

\begin{remark}
Compared to the SOTA method DiNNO \cite{dinno}, which employs the C-ADMM approach for multi-robot systems, the proposed mirror descent based algorithm, {\alg}, exploits similarities between robots to achieve communication-efficient convergence while preserving robustness against heterogeneous data through gradient tracking.
Meanwhile, unlike SONATA \cite{sonata} and Network-DANE \cite{li2020communication}, {\alg} reduces computational costs by performing a few stochastic gradient descent steps to obtain an inexact solution to the subproblem \eqref{Eq_subproblem_3} rather than solving it exactly. More importantly, our experiments demonstrate that an inexact solution with a proper choice of the number of local steps can enhance the model's generalization performance in heterogeneous data scenarios (see Fig.~\ref{fig:local_step}, where a larger $T$ indicates a more accurate solution). This may be due to the fact that overly exact solutions tend to overfit each node to its respective local optimum, leading to inconsistencies that degrade algorithm performance (c.f., $T=20$ in Fig.~\ref{fig:local_step}), especially in non-IID settings.

\end{remark}
\begin{remark}
    While existing work has provided theoretical convergence analysis for deterministic mirror descent methods applied to (strongly) convex objective functions \cite{DANE, AIDE, li2020communication, sonata}, extending such analysis to our stochastic setting without the assumption of convexity remains an open problem. This difficulty is compounded by the presence of locally inexact stochastic gradient updates. Therefore, we conduct comprehensive experiments in the next section to demonstrate the effectiveness of the proposed {\alg} and leave the convergence analysis for future work.
\end{remark}
